\documentclass[a4paper]{article}

\usepackage{INTERSPEECH2018}

\usepackage{amsmath,graphicx, color, soul, multirow}
\usepackage{amssymb}
\usepackage{subfig}
\usepackage{bm}
\usepackage{verbatim}
\usepackage{cite}
\usepackage{amssymb}
\usepackage{pifont}
\usepackage{tabularx}

\title{Improved training for online end-to-end speech recognition systems}
\name{Suyoun Kim\sthanks{The author performed the work during an internship at Microsoft Research.}$^1$, Michael L. Seltzer$^2$, Jinyu Li$^3$, Rui Zhao$^3$}
\address{
  $^1$Carnegie Mellon University\\
  $^2$Facebook\\
  $^3$Microsoft AI \& Research}
\email{suyoun@cmu.edu, mikeseltzer@fb.com, \{jinyli, ruzhao\}@microsoft.com}

\begin{document}

\maketitle
\begin{abstract}
Achieving high accuracy with end-to-end speech recognizers requires careful parameter initialization prior to training. Otherwise, the networks may fail to find a good local optimum. This is particularly true for online networks, such as unidirectional LSTMs. Currently, the best strategy to train such systems is to bootstrap the training from a tied-triphone system. However, this is time consuming, and more importantly, is impossible for languages without a high-quality pronunciation lexicon. In this work, we propose an initialization strategy that uses teacher-student learning to transfer knowledge from a large, well-trained, offline end-to-end speech recognition model to an online end-to-end model, eliminating the need for a lexicon or any other linguistic resources. We also explore curriculum learning and label smoothing and show how they can be combined with the proposed teacher-student learning for further improvements. We evaluate our methods on a Microsoft Cortana personal assistant task and show that the proposed method results in a 19\% relative improvement in word error rate compared to a randomly-initialized baseline system.

\end{abstract}
%
%
\section{Introduction}
\label{sec:intro}

Recently, several so-called end-to-end speech recognition systems have been proposed in which a neural network is trained to predict character sequences which can be converted directly to words, or even word sequences directly. Approaches to end-to-end systems typically utilize a Connectionist Temporal Classification (CTC) framework \cite{graves2006connectionist, graves2014towards, hannun2014deep, miao2015eesen, zweig2017advances, audhkhasi2018building, Li18CTCnoOOV}, an attention-based encoder-decoder framework \cite{chorowski2015attention, chan2015listen, chiu2018state}, or both \cite{kim2017joint, hori2017joint, Das18CTCAttention}. These types of models bypass much of the complexity associated with a traditional speech recognition system, in which an acoustic model predicts context-dependent phonemes, which are then combined with a pronunciation lexicon and a language model to generate hypothesized word sequences. As such, they have the potential to dramatically simplify the system building process and even outperform traditional systems as the unified modeling framework avoids the disjointed training procedure of conventional systems. 

 
Many prior studies on end-to-end models \cite{graves2014towards, miao2015eesen, chorowski2015attention, chan2015listen, amodei2016deep, kim2017joint, Li18CTCnoOOV} showed promising results with bidirectional recurrent neural networks (RNNs) that exploit access to the entire utterance. However, such models are not appropriate for online applications when the speech signal needs to be processed in a streaming manner. When there are such requirements for an end-to-end system, the most appropriate architecture is a causal model, such as a unidirectional RNN, trained with a CTC objective function. Unfortunately, when such models are trained from scratch, i.e. from randomly initialized parameters, performance is severely degraded.





As a result, many methods for initializing the parameters of online end-to-end speech recognition systems have been proposed in the literature. Most of these methods first train a tied-triphone acoustic model and then one or more intermediate models before finally having a model initialized well enough that character-based end-to-end training can be reliably performed. For phoneme-based CTC, bootstrapping the CTC models with the model trained with cross-entropy (CE) on fixed alignments has been proposed to address this initialization issue \cite{sak2015learning}. Recently, a method for training tied-triphone CTC models from scratch was proposed but it still requires a pronunciation lexicon \cite{senior2016flatstart}. Alternatively, training the models jointly using a weighted combination of the CTC and CE losses has been suggested \cite{battenberg2017reducing}. In case of the grapheme-level or word-level CTC models, it was shown that starting from phoneme-based model was essential in obtaining better performance \cite{soltau2016neural, rao2017multi, audhkhasi2017direct}. 

While these approaches significantly improve the performance of the resulting model, they are highly unsatisfying. First, it makes system development complex and time consuming, as multiple complete systems need to be trained, discarding one of the most appealing aspects of end-to-end models. Second, for lower resource languages, a high-quality pronunciation lexicon may not be available, making such initialization strategies impossible to use.

In this work, we propose a method for initializing and training a high accuracy online CTC-based speech recognition system entirely within the end-to-end framework. Our method uses teacher-student learning, curriculum learning, and label smoothing, to achieve significantly improved performance for an online end-to-end speech recognition system. We avoid any initialization that relies on a pronunciation lexicon, a senone or tied-triphone acoustic model, or any other artifacts from a traditional speech recognizer. Nevertheless, we obtain performance that approaches that of the best performing CTC systems that have been derived from tied-triphone models. 
We experimentally validate our approach on the Microsoft Cortana personal assistant task. A training set of 3,400 hours is used, demonstrating that the proposed approach is effective even when abundant training data is available.

The remainder of this paper is organized as follows. In Section \ref{sec:ctc}, we briefly review CTC-based end-to-end speech recognition. We then describe our proposed initialization and training strategies in Section \ref{sec:training} including the use of teacher-student learning, curriculum learning, and label smoothing. In Section \ref{sec:exp}, we evaluate our approach on the Microsoft Cortana task. Finally, we summarize our findings in Section \ref{sec:conclusion}.

\section{End-to-end speech recognition with character-based CTC}
\label{sec:ctc}

In this work, we use Connectionist Temporal Classification (CTC) \cite{graves2006connectionist} with character outputs for our end-to-end model, as it is well-suited to online applications. 


CTC allows the network to learn a variable-length sequence labeling problem where the input-output alignment is unknown. Given input features $\bm{x} = ( x_1,\dots, x_T )$, and the corresponding grapheme label sequence $\bm{y} = (y_1, \dots, y_U)$, CTC trains the neural network according to a maximum-probability training criterion computed over all possible alignments. The probability of the possible label sequence is modeled as being conditionally independent by the product of each label probability. The loss function of the CTC model is computed as:
    \begin{align}
    \mathcal{L}_\text{CTC} \; \triangleq & \; -\ln P(\bm{y}|\bm{x}) \approx \; -\ln  \sum_{\bm{\pi} \in \Phi} \; \prod_{t=1}^T P(k = \pi _t|\bm{x})
    \end{align} 
where $\bm{\pi}$ is a label sequence in all possible expanded CTC path alignments $\Phi$ which have the length $T$, and $P(k = \pi _t|\bm{x})$ is a label distribution at time step $t$.

CTC uses a softmax output of the network to define a label distribution $P(k | \bm{x})$, where $k$ represents each label among the $K$ graphemes and a blank symbol $\epsilon$ representing no emission of output label. Deep recurrent neural networks (RNNs) are generally used to model the distribution over labels, $P(k |\bm{x})$. Each RNN hidden layer computes the sequence $\bm{h}^l = (h_1^l, \cdots, h_T^l) $ and multiple RNN layers are stacked on top of each other. The hidden vector in the $l$-th layer, $\bm{h}^l$, is computed iteratively from $l = 1$ to $L$.
The $P(k |\bm{x})$ is defined by the softmax of the final hidden layer $h_t^L$ as follows:
\begin{align}
    \label{eq_softmax}
	P_t(k |\bm{x}) = &\; \frac{exp(h_t^L (k))}{\sum_{i=1}^{K+1} exp(h_t^L (i))} 
\end{align}
For offline applications, it is common to use bidirectional RNNs capable of modeling future context \cite{graves2013speech}.

\section{Improved character-based training for online CTC systems}
\label{sec:training}
\subsection{Teacher-student learning from offline to online models}
\label{sec:ts}

Teacher-student learning is an approach that was originally proposed to transfer knowledge from a large deep expert model (\emph{teacher}) to a smaller shallower model (\emph{student}) \cite{ li2014learning, ba2014deep, hinton2015distilling, kim2016sequence }. The student network is trained to minimize the difference between its own output distributions and those of the teacher network. This approach has been shown to be successful in several tasks including acoustic modeling \cite{ chan2015transferring }, speech enhancement \cite{ watanabe2017student }, and domain adaptation \cite{ li2017large }. Inspired by these studies, we aim to improve the accuracy of a unidirectional online model by transferring knowledge from a bidirectional offline model that does not have any latency or complexity constraints.
The recent work in \cite{takashima2018} proposed a similar teacher-student method and they use the criteria as an additional objective. Unlike their frame-level teacher-student method degrades the performance, our method improves the performance by using the teacher-student criteria in a separate training step as follows.
%

\subsubsection{BLSTM-CTC}
The first step is to build an offline end-to-end model as our \emph{teacher} model. Since there is no latency restriction, we use a deep bidirectional RNN with LSTM units (BLSTM) to predict the correct label sequence $\bm{y}$ given the entire utterance $\bm{x}$. 
Each BLSTM layer computes $\bm{h^l}$, the hidden representation in layer $l$ by combining the two hidden sequences: the forward hidden sequence $\overrightarrow{h}^l$ computed from processing the input in the forward direction, from $t=1 \rightarrow T$, and the backward hidden sequence $\overleftarrow{h}^l$ computed from processing the inputs in the backward direction from $t=T \rightarrow 1$ as follows, 
%
\begin{align}
    h_t^l = &\; W_{fw} \overrightarrow h_t^l + W_{bw} \overleftarrow h_t^l + b\\
    \overrightarrow h_t^{l+1} = &\; \text{LSTM} (h_t^{l}, \overrightarrow h_{t-1}^{l+1} )\\
    \overleftarrow h_t^{l+1} = &\; \text{LSTM} (h_t^{l}, \overleftarrow h_{t+1}^{l+1} )
\end{align}
where $W_{fw}, W_{bw}, b$ are trainable parameters, and $h_t^{0}=x_t$. 

This model is trained using the CTC objective function described in Section \ref{sec:ctc} using character labels. Because every prediction the model makes is based on observing the entire utterance (in either the forward or backward direction), the BLSTM can be reliable trained from randomly-initialized model parameters.

\subsubsection{LSTM-KL}
\label{sec:lstm_kl}
Once the BLSTM-CTC model is trained, the next step is to transfer the predictive "knowledge" of this offline model to a model that can operate in an online manner, without access to the future input frames. 
To do so, we adopt a teacher-student approach in order to train the LSTM model to minimize the Kullback-Leibler (KL) divergence between the output distributions of the offline BLSTM-CTC model and the online LSTM-CTC model.

\begin{figure}[t]
\centering
\begin{minipage}[b]{1.0\linewidth}
  \centerline{\includegraphics[width=3.5in]{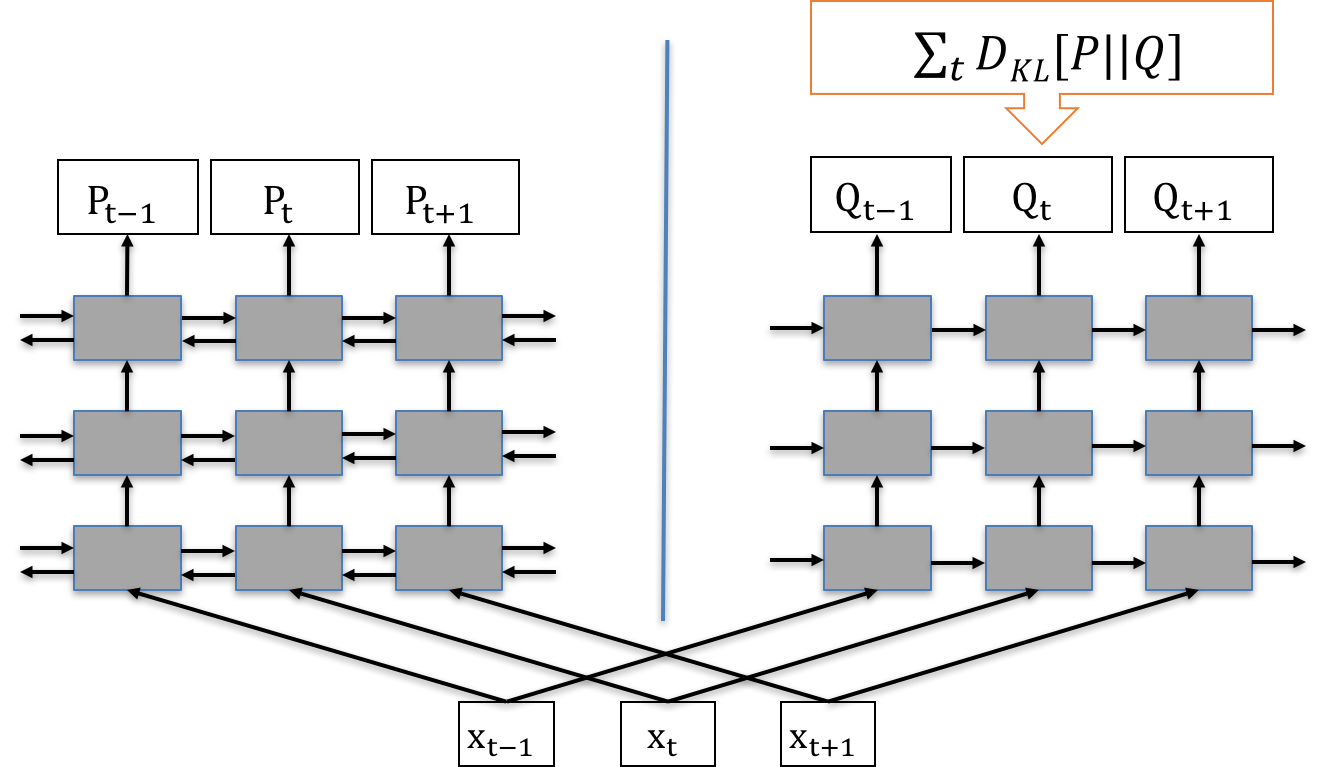}}
\end{minipage}
\caption{Our proposed teacher-student learning for online CTC models. } 
\label{fig:arch}
\end{figure}

Let $\theta_\text{BLSTM}$ be the optimized parameters of BLSTM-CTC (\emph{teacher}) , and $P_t(k | \bm{x} ; \theta_\text{BLSTM})$ be the output distribution at time step $t$ generated from the BLSTM-CTC \emph{teacher} model. Let $Q_t(k | \bm{x} ; \theta_\text{LSTM})$ be the output distribution at time step $t$ generated from LSTM-KL \emph{student} model. The goal is to find the parameters $\theta_\text{LSTM}$ that minimizes the KL divergence $D_{KL}$ between these distributions, 
\begin{align}
	D_{KL} (P_t \; || \; Q_t) =& \; \sum_{k=1}^{K+1} P_t(k | \bm{x}; \theta_\text{BLSTM})\; \ln \frac{P_t(k | \bm{x}; \theta_\text{BLSTM})}{Q_t(k | \bm{x}; \theta_\text{LSTM})} \\
    \label{eq_kl}
    =& H(P_t, Q_t) - H(P_t)
\end{align}
where $k \in \{1, \dots, K, \epsilon\} $ are the CTC labels including blank, $H(P_t, Q_t)$ is the cross entropy (CE) term and $H(P_t)$ is the entropy term. We can ignore the entropy term, $H(P_t)$, since its gradient is zero with respect to $\theta_\text{LSTM}$. Thus, minimizing the KL divergence is equivalent to the CE between the two distributions $P_t$ and $Q_t$:
\begin{align}
	\mathcal{L}_\text{KL} \triangleq & \; \sum_{t=1}^T H(P_t, Q_t) \\
	= & - \sum_{t=1}^T \sum_i P_t(k | \bm{x}; \theta_\text{BLSTM}) \ln Q_t(k | \bm{x}; \theta_\text{LSTM})
\end{align}
%
%
Figure \ref{fig:arch} illustrates teacher-student learning between the BLSTM-CTC model and the LSTM-KL model.

\subsubsection{LSTM-CTC}
The last step in the training process is to optimize the online LSTM-KL model using the CTC objective function, as described in Section \ref{sec:ctc}
\begin{align}
    \mathcal{L}_\text{CTC} (\theta_\text{LSTM}) \triangleq & \; -\ln P(\bm{y} | \bm{x};\theta_\text{LSTM})
\end{align}
In the remainder of this section, we describe two enhancements that can further improve CTC training. 

\subsection{Curriculum learning}
\label{sec:cl}
Curriculum learning has been proposed as a means of improving training stability, particularly in the early stages of learning. It was proposed to address the challenges of training deep neural networks under non-convex training criteria \cite{bengio2009curriculum}. The main idea is to present the network a ``curriculum" of tasks, starting with simple tasks and gradually increasing the task difficulty until the primary task is presented to the model. This approach is motivated by the observation that humans learn better with the examples organized in a meaningful order from simple to complex. Many prior studies \cite{mikolov2011strategies, zaremba2014learning, amodei2016deep} have shown improved performance as well as learning speed in various tasks, including language modeling, task memorization, and speech recognition. Specifically, in \cite{amodei2016deep}, they organized the training set in increasing order of the length and showed improved performance. 

In this work, we explore two different curriculum strategies. First, we train the network on subset of the training set that consists of shorter length utterances, similar to prior study \cite{amodei2016deep}. Because CTC training considers all possible sequences using the forward-backward algorithm, shorter utterances which have fewer possible paths are simpler to learn. After the network is able to learn the subset of short utterances, 
the full training set is introduced and training continues on the complete training set. 

We also propose a new curriculum, where we reduce the number of categories, and simplify the classification to only four symbols: vowel, consonant, space, and blank. As the simpler classification task is learned, the full character-based label set is restored and training proceeds.

\subsection{Label smoothing}
\label{sec:ur}
Label smoothing is a general way to improve generalization by adding label noise, which has the effect of penalizing low-entropy output distributions, i.e. overly confident predictions. The authors in \cite{pereyra2017regularizing} showed the effectiveness of such a strategy on several well-known benchmark tasks, including image classification, language modeling, machine translation, and speech recognition. However, they have only explored the effectiveness of label smoothing in an encoder-decoder framework \cite{chorowski2015attention, chan2015listen}. 

We observed that the label distribution from the CTC models is primarily dominated by the blank symbol. Motivated by this observation, we penalize low-entropy output distributions, similar to prior study \cite{pereyra2017regularizing}.
To do so, we add a regularization term to the CTC objective function which consists of the KL divergence between the network's predicted distribution $P$ and a uniform distribution $U$ over labels. 
\begin{align}
	\label{eq:ur}
	\mathcal{L}(\theta_{\text{online}}) \triangleq & \; (1 - \alpha) \mathcal{L}_\text{CTC} + \alpha \sum_{t=1}^T \; D_{KL}(P_t || U)
\end{align}
where $\alpha$ is tunable parameter for balancing the weight regularization term and CTC loss. 

\section{Experiments}
\label{sec:exp}

\subsection{Datasets}

We investigated the performance of our proposed training strategies on Microsoft's U.S. English Cortana personal assistant task. The training set has approximately 3,400 hours of utterances, the validation set has 10 hours of utterances, and the test set has 10 hours of utterances. As acoustic input features, we used 80-dimensional log mel filterbank coefficients extracted from 25~ms frames of audio every 10~ms. We employed a frame-skipping approach \cite{sak2015fast}, where three consecutive frames are concatenated to obtain a 240-dimensional feature vector. None of our experiments used a pronunciation lexicon except when we compared to a system initialized with a tied-triphone model.
Following \cite{zweig2017advances}, we used a label symbol inventory consisting of the individual characters and their double-letter units. An initial capitalized letter rather than a space symbol was used to indicate word boundaries. This results in 81 distinct labels derived from  26 letters.  

\subsection{Training and Decoding}

Our online LSTM-CTC model was a 5-layer LSTM \cite{hochreiter1997long, graves2013hybrid} with 1024 cells in each layer. Each LSTM layer has a linear projection to 512 dimensions. The offline BLSTM-CTC was a 5-layer network with 1024 cells in each direction, forward and backward. The hidden layer outputs in both directions were projected down to a 512-dimensional vector. When the models were randomly initialized, all the weights of our models were initialized with a uniform distribution in the range [-0.05, 0.05]. Parameters were trained using stochastic gradient descent with momentum. We use an initial learning rate of 0.0001 per sample. After each epoch, the criterion of the development set is evaluated, and if performance has degraded, the learning rate was decreased by a factor of 0.7. 

For the decoding, the most likely sequence of characters was generated by the model in a greedy manner. The final output sequence was then obtained by removing any blank symbols or repetitions of characters from the output and replacing any capital letter with a space and its lowercase counterpart. Note that we did not use any lexicon or language models.

\subsection{Results}


We first compared the performance of an online LSTM model trained from random initialization to an LSTM trained by bootstrapping from a tied-triphone model. The tied-triphone initialization procedure was as follows. First, the LSTM was trained using cross-entropy computed on frame-level tied-triphone labels. Then, the model was retrained using CTC on tied-triphone outputs. Finally, the output layer was removed and replaced with character labels and then trained using a character-based CTC criterion. 

As shown in Table \ref{tab:result1}, a randomly initialized system obtains a WER of 41.0\%, while a tied-triphone-initialized system performs significantly better, with a WER of 30.8\%, a 25\% relative improvement. This difference in performance demonstrates that proper initialization is critical in training online end-to-end systems. Our goal is to close this gap in performance without relying on any of the linguistic resources required to build a tied-triphone acoustic model. 

We next evaluated the three proposed training strategies individually, and all three resulted in substantial improvements in WER over the randomly-initialized baseline system. Curriculum learning, label smoothing, and teacher-student model initialization provided 7.8\%, 13.7\%, and 12.2\% relative improvement in WER, respectively. The label smoothing technique showed largest improvement when a value of $\alpha = 0.05$ was used in Equation (\ref{eq:ur}). We also found that curriculum that focused on a simplified label set also improved performance but did not work as effectively as one that focused on short utterances, so the results are not shown.

When we combined the two top performing strategies, teacher-student initialization and label smoothing, additional improvement is obtained, showing a 33.9 \% WER. In this configuration, we first trained an LSTM-KL model, as described in Section \ref{sec:lstm_kl}, and then used it as an initial model for training with the CTC loss and the uniform distribution regularization. Lastly, we evaluated the combination of all three training strategies. This method performed best and obtained 33.2 \% in WER, a 19\% relative improvement over a randomly-initialized LSTM model. For reference, the table also shows the performance that can be obtained using an offline bidirectional model. Interestingly, even the performance of the BLSTM is improved considerably from label smoothing. 

Another way to interpret these results is by considering how much the gap in performance between the randomly-initialized system and the tied-triphone-initialized system was narrowed by the proposed approach. There is a difference of 10.2\% absolute between the randomly-initialized system and the tied-triphone-initialized system and the proposed approach reduces this gap to 2.4\%. Thus, the proposed approach has closed the gap by over 75\% in a way that does not rely on a pronunciation lexicon. 

\begin{table}[t]
\caption{ WER on test set of models with various training strategies, curriculum learning (CL), label smoothing regularization (LS), and teacher-student approach (TS). None of our experiments used any language model or lexicon information.}
  \vspace{-0.2cm}
\label{tab:result1}
\begin{center}
\begin{tabular}{l | c }
\hline
Training Strategy            & WER(\%) w/o LM \\
\hline
LSTM + random initialization & 41.0    \\
LSTM + tied-triphone pre-training     & 30.8    \\
\hline
LSTM + CL                    & 37.8    \\
LSTM + LS                   & 35.4    \\
LSTM + TS                    & 36.0    \\
LSTM + TS + LS               & 33.9    \\
LSTM + TS + CL + LS          & 33.2    \\
\hline
BLSTM						& 27.8 \\
BLSTM + LS                   & 25.5    \\
\hline
\end{tabular}
\end{center}
\end{table}

\begin{figure}[!h]
\begin{minipage}[b]{1.0\linewidth}
  \centering
  \centerline{\includegraphics[width=8.5cm]{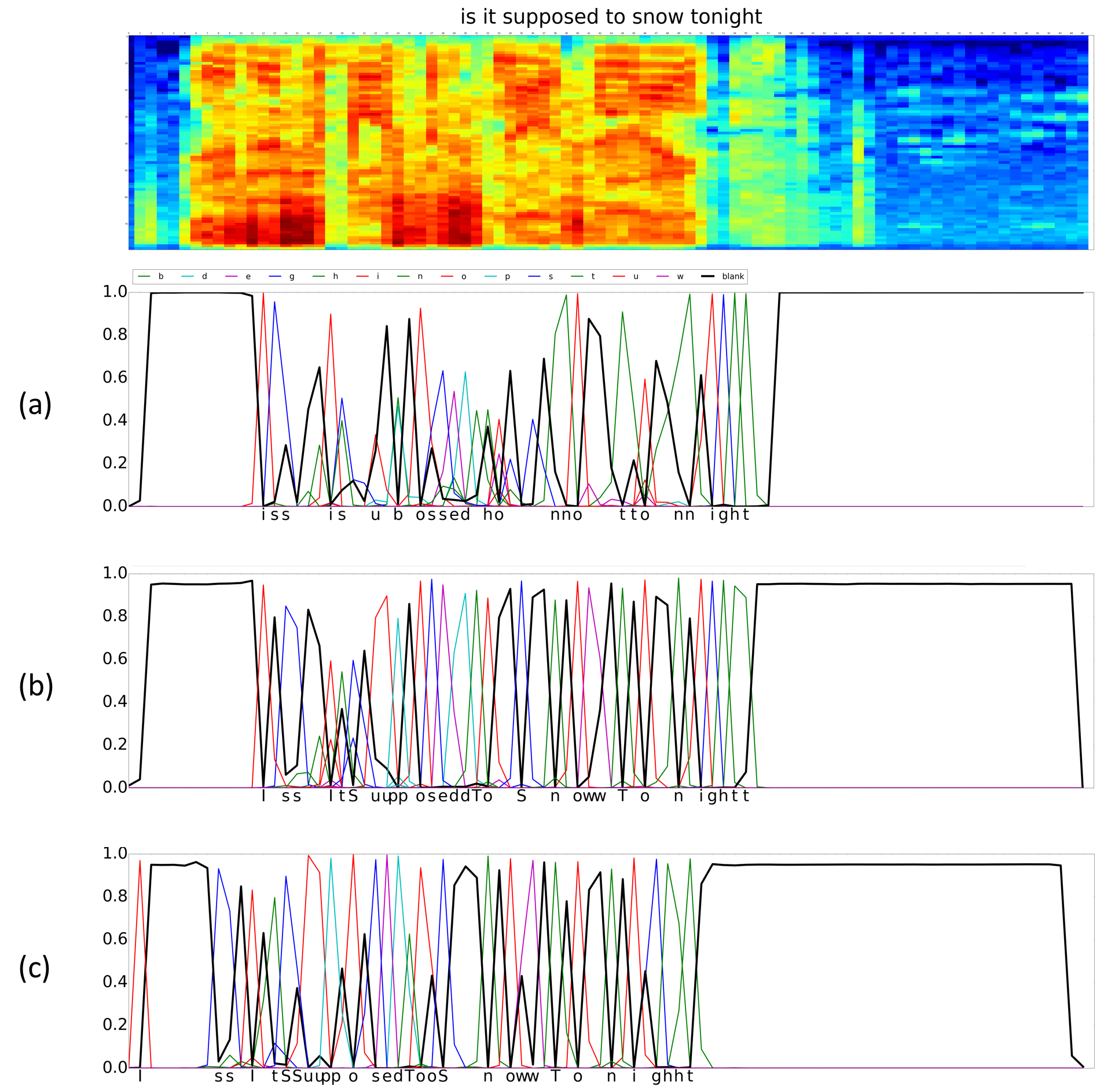}}
\end{minipage}
\caption{Comparison of the CTC label probabilities over the input frames of the utterance "Is it supposed to snow tonight". for various models: a) an LSTM with random initialization, b) our proposed model with teacher-student learning, LSTM + TS + CL + LS,  and c) an offline model BLSTM + LS. 
}
\label{fig:prob}
\end{figure}

Figure \ref{fig:prob} shows the log-mel feature vectors for an utterance and the corresponding character probabilities using models trained with different training strategies. The black line shows the probability of the blank symbol and the other colors correspond to other letters. We manually chose an utterance that was recognized incorrectly by the randomly-initialized LSTM but correctly recognized by both the LSTM trained with the proposed approach and a BLSTM trained with label smoothing. 

The figure shows that an LSTM initialized with teacher-student training has consistently higher probability on the most likely output character compared to the randomly-initialized where several of the most-likely characters only have a posterior probability of 0.6 or less. It is also interesting to note the differences in spike locations between the online models and the offline BLSTM models. The BLSTM models can often have spikes that precede the acoustic observations of the corresponding letter as a result of the model processing information in the forward and backward directions simultaneously.

\section{CONCLUSION}
\label{sec:conclusion}

We proposed a method to improve the accuracy for an online end-to-end speech recognizer via teacher-student learning. We transfer knowledge from a large, well-trained offline BLSTM model to an online LSTM model, by minimizing the KL divergence between the label distributions of the online model and the offline model. Because our method does not require bootstrapping from a tied-triphone system, it simplifies the training procedure and is beneficial for the languages where a pronunciation lexicon may be unavailable. We also found that our method can be easily combined with curriculum learning strategies and label smoothing. Our model was shown to outperform models that are trained from random initialization, and approach the performance of models bootstrapped from tied-triphone systems.

\bibliographystyle{IEEEtran}

\bibliography{refs}


\end{document}